# Language Models for Handwritten Short Message Services

Emmanuel Prochasson
*LINA – FRE CNRS 2729*
*Université de Nantes*
*Emmanuel.Prochasson@univ-nantes.fr*

Christian Viard-Gaudin
*IRCCyN UMR CNRS 6597*
*Université de Nantes*
*Christian.Viard-Gaudin@univ-nantes.fr*

Emmanuel Morin
*LINA – FRE CNRS 2729*
*Université de Nantes*
*Emmanuel.Morin@univ-nantes.fr*

**Abstract**

*Handwriting is an alternative method for entering texts composing Short Message Services. However, a whole new language features the texts which are produced. They include for instance abbreviations and other consonantal writing which sprung up for time saving and fashion. We have collected and processed a significant number of such handwriting SMS, and used various strategies to tackle this challenging area of handwriting recognition. We proposed to study more specifically three different phenomena: consonant skeleton, rebus, and phonetic writing. For each of them, we compare the rough results produced by a standard recognition system with those obtained when using a specific language model.*

## 1. Introduction

SMS (Short Message Service) has achieved huge success in the wireless world. It is a technology that enables the sending and receiving of messages between mobile phones. As suggested by the name "Short Message Service", the data that can be held by an SMS message is very limited. One SMS message can contain at most 140 bytes (1,120 bits) of data, so one SMS message can contain up to:

- 160 characters if 7-bit character encoding is used. (7-bit character encoding is suitable for encoding Latin characters like English alphabets.)
- 70 characters if 16-bit Unicode UCS2 character encoding is used. (SMS text messages containing non-Latin characters such as Chinese characters should use 16-bit character encoding.)

Person-to-person text messaging is the most commonly used SMS application and it is what the SMS technology was originally designed for. In these kind of text messaging applications, a mobile user types an SMS text message using the keypad of his/her mobile phone, then he/she enters the mobile phone number of the recipient and finally sends the text message out.

However, the small phone keypad and the limited message lengths caused a number of adaptations of spelling, as in the phrase "txt msg", or use of CamelCase, such as in "ThisIsVeryCool". Users aim to use the least number of characters needed to transmit a comprehensible message. Hence, punctuation and grammar are largely ignored [5].

To circumvent the bottleneck of the keyboard entry, two quite different strategies are encountered: one is to assist the user with optimized predictive text entry solutions ([1], [2]), which consists in some form of disambiguation to determine which letter, among the three or four letters shared by the same key, is intended by the writer (see [1] for complete references). Another is to replace the keyboard by handwriting input, using either a stylus and a screen, or a digital pen and paper solution connected to the GSM phones. Our goal here is to improve recognition of handwritten message to allow them to be sent like normal text SMS.

It has been proved that language models allow to increase significantly the recognition rate of handwriting systems [3]. They allow minimizing the error recognition rate by taking into account the context in order to disambiguate poorly written texts. Two approaches are likely to be implemented. One is based on structural models specifically designed by linguistic experts, while the other approach relies on some statistics computed on large written text corpora. One example of the latter being the well-known n-gram models, which work either at the character or at the word levels.

In this paper, we have identified three different phenomena that alter SMS texts, and we propose for each of them a specific adaptation of the handwriting recognition engine.

## 2. SMS language

For both technical reasons – limited length of the text and multiple taps of the key, and sociological reasons – short messages are particularly popular amongst teens, several phenomena affect SMS texts when compared to standard written productions. A few are listed below.

## 2.1. Rebus style

Rebus style writing is characterized by using a single letter or digit to replace a whole syllable or word. Examples are:

–*be* → *b* ; *you* → *u* ; *are* → *r* (single letter replace whole word) ;
–*ate* → *8* ; *for* → *4* ; *to, too* → *2* (single digit replace a whole word) ;
–*skate* → *sk8* ; *later* → *l8er* ; *before* → *b4* (letter or digit replace a whole syllable within word) ;

Using only rebus style, one can easily construct a whole phrase, for example "*c u l8er!*" (*see you later!*).

## 2.2. Consonant Skeleton style

Consonant Skeleton style is characterized by the withdrawal of most of the vowels of a word, leaving only the "consonant skeleton" of the word. Obvious examples are "text → txt", "people → ppl". This transformation can keep some vowels and should not necessarily keep all consonants (nasal consonants, when combined to nasal vowels are removed most of the time).

## 2.3. Phonetic style

This phenomenon follows few morphological rules, except that transformed text, when read, should keep roughly the same pronunciation than the genuine one, for example: "*giv me som luv*" (*"give me some love"*).

Although this phenomena is pretty rare in English, it is very common in French, since French word often contains mute letters that can be removed without modifying the pronunciation of the word.

## 2.4. Mixed styles

Combinations of the above can shorten a single or multiple words. In addition, punctuation marks could be removed ; only period and exclamation marks are generally used. The space and capital letter are often omitted after a period. Whole words may be omitted, especially articles.

Other transcriptions of slang or dialect terms can be used if shorter than the original words, as in "*cos*" (standing for "*because*").

Combining the above "techniques" can shorten whole sentences. A few more examples are:

–*Are you going to the pub tonight?*
–**ru goin pub 2nyt**

–*Hi mate. Are you okay? I am sorry that I forgot to call you last night. Why don't we go and see a film tomorrow?* (120 characters)
–**hi m8 u k?-sry i 4gt 2 cal u lst nyt-y dnt we go c film 2moz** (60 characters)

Similar examples can be found in the SMS corpora provided by [1]. Note that the transformation leads to new way of writing known words, such word is subsequently called a "neography".

This field of research has not been to the best of our knowledge yet investigated by the handwriting recognition community. However, it is a very challenging application where many problems have to be solved. To begin with, there is no database available as handwritten samples, although it exists a few studies involving electronic corpora, such as in [1] in English, and [4] in French.

## 3. Handwritten Short Messages corpora

To assess the performances of the proposed algorithm, we have collected a database of French Handwritten Short Messages (HSM). It contains a total of 1,221 HSM, representing 38,462 characters and 11,600 words. The number of writers is 150, each having written from 6 to 8 HSM. As a device to collect the ink, we have used a digital pen and paper solution, which allows to communicate with phones using blue-tooth connection. Two handwriting styles have been asked: isolated characters HSM, with boxed form, and unconstrained HSM, with only a baseline. Concerning the message itself, either it was imposed to the writer, or conversely, we asked the writer to propose his/her own HSM. Table 1 synthesizes information about this corpora.

|            | Boxed Handwriting | Cursive Handwriting |
|------------|-------------------|---------------------|
| Given text | 177               | 174                 |
| Free text  | 493               | 477                 |
| Total      | 670               | 551                 |

**Table 1: The HSM corpora**

Fig. 1 shows an example HSM. Text written is "*Bizoo a2m1*", translated in standard French: "*Bisous, à demain*" ("*Kiss, see you tomorrow*"). In addition to the on-line information, for each sample, we know the label at the HSM level.

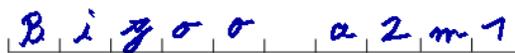

**Fig. 1: a sample HSM – boxed handwriting**

It is cautious to note that using an electronic pen instead of a small cellular phone keypad is likely to modify the structure of input text. Indeed, as it is easier to write with a pen, there might be no more needs for complex abbreviations. However, we can not asses the importance of this evolution since this particular use will need time to get installed (who could have predicted the emergence of SMS style language ?).

## 4. Baseline recognition results

### 4.1. Recognition engine

We have used a high quality industrial Software Development Kit (SDK) for on-line handwriting recognition. This product allows to use and combine several Language Models but also to build new one, using stochastic Regular Expression or lexicons (see part 5). The contribution of this paper will be to investigate which kind of additional resources could be developed to extend the performances of the recognizer to process HSM.

Two standard French language models (Linguistic Knowledge – LK) are provided with recognition engine [6], they are:

–*LK-text*: for "correctly" written text, it contains a very large lexicon of common French words, a French language model (mostly based on n-grams of words) and however, it enables to recognize out-of-lexicon words.

–*LK-free*: for un-characterized French text. No lexicon is used, but a basic language model, which performs only at the letter level to allow recognition of unknown words.

Fig. 2 illustrates the global recognition process. As shown, multiple resources can be combined for recognition, but one can also choose not to use any linguistic knowledge.

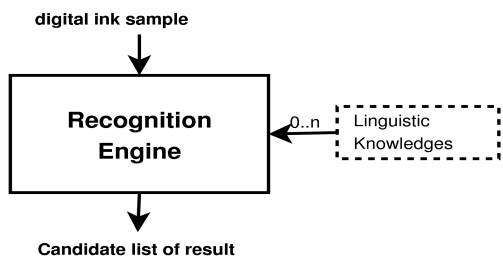

**Fig. 2: Recognition process**

Candidate list of results provided by recognition engine is sorted according to a recognition score. We have always considered only the first candidate. It is also worth to note that the recognition engine is already trained, and that neither the HSM database, nor the writers of this database have been involved in the training stage.

### 4.2. Performance assessment

As the database is labeled at the HSM level, measurement of the recognition rate is not straightforward. Indeed, having a recognition rate at this level would not be very meaningful, it is more significant to work at the character level. Consequently, we have used a recognition rate derived from the Levenshtein distance ($D$) computed between the two strings which are the first candidate produced by the recognition engine and the corresponding label of a given HSM. The edit costs have been set to 1 for deletion and substitution, and to 0 for the insertion operation, in order to not penalize twice an over segmentation problem, which most of the time lead to a substitution and an insertion operation.

The recognition rate ($RR$) at the character level will be computed as follows:

$$RR = 100 \times (\#label - D)/\#label$$

It turns out that if all characters of the HSM are recognized, then $D = 0$, and $RR = 100\%$. Conversely, when none are recognized, then $D = \#label$, and $RR = 0\%$. This formulation allows to bound the recognition rate between *0* and *100*, but it does not penalize additional characters in the recognized string.

Table 2 illustrates this computation on a basic example.

| Label | bjr | #label = 3 |
|---|---|---|
| Recognition result | loj.t | RR = (3-2)/3=0.33 |

**Table 2: example of bad recognition, with insertion**

### 4.3. Baseline results

Table 3 presents recognition results with different linguistic knowledges, for both unconstrained-handwritten and boxed-handwritten messages.

| | Using no additional LK (cursive / boxed) | Using optimal LK (cursive / boxed) |
|---|---|---|
| No LK | 87% / 94% | 96% / 96% |
| lk-text | 84% / 90% | 95% / 96% |
| lk-free | 88% / 95% | 88% / 95% |

**Table 3: Recognition quality for cursive/boxed handwriting**

First column corresponds to normal use of the recognition system with either no language model, *LK-text* or *LK-free* resources, while in second column an additional optimal lexicon, which consists of the exact list of words contained in labels, is added to theses resources. Thus, these later results give an upper bound for the recognition rate.

Clearly, with such kind of texts using *LK-text*, which is the standard model for French text, is not recommended. Conversely, the *LK-free* resource outperforms slightly a recognition system without any language model.

In the next section, we would like to define specific language resources to model the main phenomena that alter texts, so that they can be recognized more easily by the recognition system.

## 5. Modeling specific SMS languages

In this part, we present every language models developed for each phenomena.

### 5.1. Consonantal skeleton

Consonantal skeletons, as described in section 2.2, can be defined either extensively with an appropriate lexicon generated from the standard lexicon using appropriate rules or with a more formal model based on a stochastic regular expression. We have used later approach, and based on an analysis of the observed phenomena, we come to the model proposed in Fig. 3.

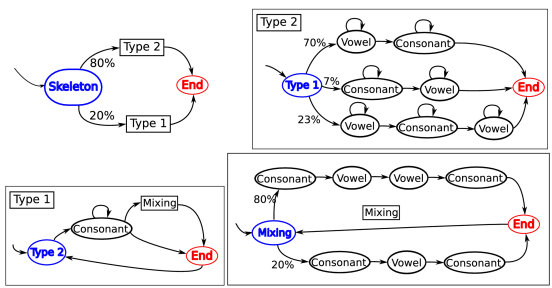

**Fig 3: Stochastic Finite State Automata for Consonant Skeleton**

This Regular Expression describes the following constraints, resulting from observations on frequent French words:
−Most of skeletons (80%) are made of consonants, where a few vowels can be kept, for example "*bonjour*" → "*bjour*".
−Some skeletons (20%) begin (70%) or end (7%) with a vowel (or both: 23%).

We also tried to build a lexicon using a few rules to transformed a word into its consonant skeleton form. We obtained 3,244 words in the lexicon, covering only 22% of the skeleton sub-corpora, due to the nature of the initial corpora (we used French newspaper "*Le Monde*", where words such as "*salut*" – "*hello*", very frequent in the consonant skeleton corpora are obviously rare). Using the lexicon introduces noise and in no way helps improving the recognition quality. We therefore choose not to use it any further. That does not mean that lexicon can not be used, but a more adapted corpora needs to be chosen.

Recognition results concerning consonantal skeleton are synthesized in part 5.4.

### 5.2. Rebus

We proceed with the same kind of approach for modeling Rebuses. Another Stochastic Regular Expression has been defined. It follows the general constraints:
−Half of rebuses are singletons, and some singletons are frequent (such as "*2*", used to replace most frequent French bi-gram "*de*"), whereas others are really rare ("*4*" – "*quatre*" alone can not replace any frequently used syllable in French).
−Half of rebuses mix letters and numbers (and few symbols like "+" and "-") ; and there is no sequence of numbers.

Results for recognition using this Regular Expression are detailed in part 5.4.

### 5.3. Phonetic writing

Phonetic writing is the hardest transformation to model. Indeed, from a morphological point of view, phonetisation does not present any specific structure, conversely to rebus and consonant skeleton writing. However the process of phonetisation is quite simple: most mute letters can be easily identified (for example: "*e*" after a vowel at the end of a French word) and a lot of simplifications can be done automatically. Thus, it is possible to define a set of rules that allows to generate homophonic words from a given word. Some of these rules are:
−withdrawal of mute [*e*] and mute consonants (mostly at the end of words – typically [s], mark of the plural) ;
−withdrawal of double consonant ([*ll*] → [*l*]) ;
−withdrawal of [*h*], when not combined with <*c, p, s*> ;
−miscellaneous transformations ([*au*] → [*o*] ; [*qu, c*] → [*k*] ; [*ç*] → [*c*] ; [*ai, é, è, ais, ait*] → [*é, è*]) ;

Every possible transformations are applied to each words and to the results of the transformation. For example, starting with the word "*musique*" we generated the list "*muzique, musiqu, musike, musike, muziqu, muzike, musik, muzik*"

We have selected the most frequent French words, from a corpora based on the newspaper "*Le Monde*", and then have automatically applied the above rules to produce a new lexicon that contains all the homophonic words. In this experiment, 3,171 homophonic words from most frequent 1,200 French words have been used.

### 5.4. Results

We have manually classified each HSM in one of the four categories: consonant skeleton, rebus, phonetic style, and others. 54 consonant skeleton-style (151 char.) ; 96 rebus-like (222 char.) ; and 91 phonetic-style (327 char.) samples were extracted from Boxed-HSM corpora. Then we applied the proposed additional resources to the recognition system and compare with the baseline results (lower bound: best result without any modification of the initial LK provided), and with the optimal lexicon (upper bound). Table 4 synthesizes results for all three phenomena.

|  | Lower Bound | Proposed LK | Upper Bound |
|---|---|---|---|
| Consonant Skeleton Style | 94.7% | 98.0% | 100% |
| Rebus Style | 92.6% | 92.6% | 94.6% |
| Phonetic Style | 94.1% | 94.1% | 99.3% |

**Table 4: Recognition Result using proposed language models**

Those results require explanations. In the case of the consonant skeleton style, the quality is improved (better said: nearly 38% of initial errors get corrected). In the cases of Rebus Style and Phonetic Style, improvement is not directly visible. Nevertheless, more detailed results (see table 5) show that proposed language models greatly improved initial *RR* when combined with standard French language models (*LK-text*).

|  | LK-text | LK-text + proposed LK | LK-text + optimal |
|---|---|---|---|
| Consonant Skeleton Style | 66.2% | 98.0% | 98.0% |
| Rebus Style | 69.1% | 92.1% | 94.6% |
| Phonetic Style | 75.2% | 90.5% | 99.0% |

**Table 5: Recognition results using lk-text and proposed LK**

Table 5 clearly indicates that proposed models brings a lot of useful information to a poorly adapted initial LK. Nevertheless, those result are yet worst than results found using proposed LK only. The *LK-text* LK is over constrained and in all case inadequate for HSM recognition. Combining *LK-text* with proposed LK helps correcting nearly 62% of initial errors in the case of Phonetic Style, and 75% of initial errors in the case of rebus style.

### 6. Conclusion

Although SMS are now very common, only a few studies has been conducted around this new way of communication. Our work, in the field of this study, was first to describe and understand precisely each phenomena to be able to characterize them within the recognition engine. This is an arduous task since those phenomena are essentially creative and un-constrainted, and bringing too much or too less information to the recognition engine would just lower the quality of the recognition results.

We introduced here some successful propositions for characterizing isolated, known transformations. A lot of work is still to be done: some transformation has not been treated yet and proceeding combinations of those transformations as shown in part 2.4. is far from being an easy task. Finally, to be able to perform handwritten SMS recognition, some works need to be done to finely tune and combine introduced LK in order to build a global SMS Linguistic Knowledge. Indeed, we hereby worked on isolated phenomena, but we would have needed more control onto the recognition engine. We could only use it as a black box, with LK combinations, and were not able to tune it up finely enough to integrate efficiently our work.